\documentclass[10 pt, conference]{ieeeconf}  

\IEEEoverridecommandlockouts                              

\usepackage{amssymb,url,amsmath,bm}
\usepackage{algorithm}
\usepackage{graphicx}
\usepackage{epstopdf}
\usepackage{textcomp}
\usepackage{color}
\usepackage{ifpdf}
\usepackage{url}
\usepackage{enumerate}
\usepackage{array}
\usepackage{flushend}
\usepackage{multirow,booktabs}
\usepackage[pdfstartview=]{hyperref}
\usepackage[export]{adjustbox} 
\usepackage{caption}
\usepackage{subcaption}
\usepackage{mdwlist}
\usepackage[utf8]{inputenc}
\usepackage{todonotes}

\usepackage{tcolorbox}
\newtcolorbox[auto counter]{algorithmbox}[2][]{colback=red!5!white,colframe=red!75!black,fonttitle=\bfseries, title=\alg\thetcbcounter: #2,#1}

\newcommand{\alg}[1]{Algorithm~\ref{#1}}

\newcommand{\eq}[1]{Eq.~\ref{#1}}
\newcommand{\fig}[1]{Fig.~\ref{#1}}

\newcommand{\sect}[1]{Sec.~\ref{#1}}

\usepackage{cite}

\title{Dyna-T: Dyna-Q and Upper Confidence Bounds Applied to Trees}

\author{Tarek Faycal$^{1}$ and Claudio Zito$^{2\ast}$
\thanks{$^{1}$IRLab, School of Computer Science, University of Birmingham, United Kingdom}%
\thanks{$^{2}$Technology Innovation Institute, Abu Dhabi, UAE}
\thanks{$^{\ast}$Correspondent author
{\tt\small Claudio.Zito@tii.ae}}
}
\date{}

\begin{document}

\maketitle

\begin{abstract}
In this work we present a preliminary investigation of a novel algorithm called Dyna-T. In reinforcement learning (RL) a planning agent has its own representation of the environment as a model. To discover an optimal policy to interact with the environment, the agent collects experience in a trial and error fashion. Experience can be used for learning a better model or improve directly the value function and policy. Typically separated, Dyna-Q is an hybrid approach which, at each iteration, exploits the real experience to update the model as well as the value function, while planning its action using simulated data from its model. However, the planning process is computationally expensive and strongly depends on the dimensionality of the state-action space. We propose to build a Upper Confidence Tree (UCT) on the simulated experience and search for the best action to be selected during the on-line learning process. We prove the effectiveness of our proposed method on a set of preliminary tests on three testbed environments from Open AI. In contrast to Dyna-Q, Dyna-T outperforms state-of-the-art RL agents in the stochastic environments by choosing a more robust action selection strategy.    
\end{abstract}

\section{INTRODUCTION}\label{sec:introduction}

At the core of reinforcement learning (RL) there is the problem of how to update an approximate value function such that a policy is found. Some methods require a model of the environment to refine a policy by interacting with simulated experience from the modelled environment. These methods are typically referred to as \textit{planning} methods. Other methods learns directly from experience without the need of a model, and they are referred to as \textit{direct RL} methods. Within planning agents, it is still possible to use the experience collected by interacting with the environments to improve the model, called \textit{model-learning} or \textit{indirect RL}. Both direct and indirect approaches present advantages and disadvantages. Indirect methods tend to exploit better a limited amount of experience, thanks to their models, and thus achieve better policies with few interactions. On the other hand, direct methods are typically simpler and do not suffer any bias from the design of the model. 

Dyna-Q is a form of hybrid method that uses the real experience generated by acting on the environment for learning a model as well as a value function. The value function is then refined by simulated experience generated by the model. Acting, direct RL, model learning, and planning happen in the Dyna agent at each interaction. The planning process is usually the computationally intensive phase of the algorithm which strongly depends on the dimensionality of the state-action space, as well as the complexity of the problem.

In this paper, we propose to improve the planning efficiency of Dyna-Q by constructing a upper bound confidence tree (UCT) on sampled simulated experience. We call this new algorithm Dyna-T.

UCT is a state-of-the-art approach to solving large state-space planning problems. The algorithm estimates the utility of a state by building and searching on a tree of simulated experience via a procedure called \textit{rollout}. The key aspect is that not all the branches of the tree are equally sampled so to dedicate the most of the resources versus most promising paths. UCT achieves that by treating each node of the tree as a multi-armed bandit problem. Since random sampling is used to discover nodes with a good return, UCT could take a long time to converge to good performance. Many enhancements have been proposed to address this problem by either (i) tweaking the action selection formula (e.g. which arm is pulled), or (ii) designing better-informed rollout policies. Our proposed approach fall in the second category. However, different than many of the previous attempts, we do not based our rollout exploration on heuristic-based policies, but on a model-based RL policy provided by the Dyna-Q algorithm.



\section{BACKGROUND}\label{sec:background}

In this section, we briefly introduce the basic concepts about the nature of RL and the specifics of designing agents for this machine learning paradigm. We will focus on state-of-the-art approaches relative to this work and highlight some of their strengths and drawbacks. A comprehensive review of this topic is out of scope and we redirect the interested reader to a few available sources, such as \cite{russell2016artificial, sutton1998reinforcement, bellman1957markovian}.  

\subsection{Reinforcement Learning}

A traditional reinforcement learning setting consists of two components: An agent and an environment. The agent receives an observation from the environment that contains the current state the agent is in, as well as a reward signal that describes how favorable that current state is compared to others. The agent would then map that state to an action using its policy function returns an action to the environment that is intended to alter the current state \cite{russell2016artificial}. 

One iteration of this loop is a single step in the environment. The agent will continue taking steps in the environment until a predetermined number is reached or some environment-specific termination condition is met. This interaction is termed episodic, because it has a clearly defined start and end.   

This way of learning poses a unique difficulty when compared to other branches of machine learning. The simulation of experience might be computationally expensive depending on environment complexity, added to the existing cost associated with learning. Furthermore, the environments can vary greatly making a general purpose reinforcement learning algorithm or agent difficult to construct. 

\subsection{Markov Decision Processes}

One mathematical framework that can be used to model reinforcement learning problems is a Markov Decision Process (MDP) \cite{bellman1957markovian}. An MDP consists of a set of states $S$ that describe the environment, a set of actions $A$ available to the agent, a reward function $R(s,s')$,  which returns the reward gain from moving from state $s$ to state $s'$, and a transition model $P(s'|s,a)$ that returns the probability of moving to state $s'$ given the current state $s$ and action $a$. The solution to this problem comes in the form of a policy function $\pi(s)$ that returns an action given a state. The optimal policy is the one that maximizes the rewards collected by the agent \cite{sutton1998reinforcement}. The important assumption in this framework is the process follows the Markov property, meaning the current state of the environment must contain all information required by the agent to make a decision regarding what action to take, which implies that the agent does not need access to the history of this process. 
However, many real-world applications in robotics are characterised by continuous state, action and observational spaces and/or physical interactions with the environment challenge the MDP formulation, although several attempts have been made in the past\cite{bib:zito_2016, bib:zito_w2012, bib:zito_w2013, bib:rosales_2018, bib:zito_2019}. 

\subsection{Active RL and the role of policy in exploration}

There are two approaches to learning within the MDP framework: active and passive. Passive RL evaluates a fixed policy over a certain number of trials in the environment to derive the utility value associated with each state. Utility is a concept borrowed from Economics and represents how good a state is in the context of the problem and policy being evaluated. It is computed as the expected sum of discounted rewards $U^{\pi}(s) = \mathbb{E}[\sum\limits_{t=0}^{\infty} \gamma R(s_t)]$. The $\pi$ symbol is the policy being evaluated, $\gamma\,\epsilon\,[0,1]$ is a discount factor used to control the influence of future states on the value of the current one, and $R(s_t)$ is the reward gain in state $s$ at time $t$.

Active RL adopts a different approach, the agent must interact with the environment and dynamically adjust its policy in order to arrive at an optimal one. This introduces a new problem, known as the exploration/exploitation dilemma. How much time should an agent spend exploring its environment, and when should it stop and exploit the knowledge it has gathered so far to maximize the reward? There is no single answer to this question because of the wide range of possible environments. For example, in an environment with very sparse rewards, the agent may need to spend a significant amount of time exploring before finding any reward signal. Additionally, If it does come across a solution early, how will the agent know to stop then? and how can it guarantee that there aren't any higher rewards to be gained from further exploration? \cite{sutton1998reinforcement}. This work will focus on active RL given that it is a more accurate representation of how biological agents learn, in addition to the fact that some environments cannot be simulated due to high complexity or dynamics that are not well understood.  

One of the most commonly used strategy to guarantee exploration is 
$\epsilon-greedy$. The agent always takes the best action according to the information gathered so far, but with a small probability $\epsilon$ of choosing a random action from those currently available. Presented by Sutton and Barto in 1998 \cite{sutton1998reinforcement}, it has remained one of the most commonly used strategies in state of the art research, albeit with some modifications such starting with $\epsilon = 1$ and gradually reducing it to converge on the learned policy. 

\subsection{Model-Free and Model-Based Learning}

One approach to solving MDPs is framing it as an optimization problem. Adaptive Dynamic Programming (ADP) was one of the first passive RL methods utilized. The underlying principle is that the utility values of each state obey the Bellman Equation \cite{dixit1990optimization} for a fixed policy $\pi$: 

\begin{equation}\label{eq:bellman}
    U^\pi(s)=R(s)+\gamma \sum_{s'} {P(s'|s,\pi(s))}{U^\pi(s')}
\end{equation}

Given a fixed policy, and the above condition, the agent can learn the transition model $P(s'|s,\pi(s))$ by going through a certain number of episodes and collecting statistics about state transitions and observed rewards. This information can be plugged in to \eq{eq:bellman} and solved for each state, either directly or through iterative methods, giving us a utility estimate for each state: 

\begin{equation}\label{eq:iterative_bellman}
U_{i+1}(s) \leftarrow R(s) + \sum_{s'} P(s' | s, \pi_{i}(s))U_{i}(s')
\end{equation}

Equation \ref{eq:iterative_bellman} shows a simplified iterative update to the utility where $i$ is the current iteration of the algorithm. The above equation can be used as part of a method called policy iteration \cite{russell2016artificial} that starts with a random policy and updates it after every run of utility evaluation to take the actions that maximize the final reward gathered. 

Such a method is called \textbf{model-based} because it explicitly learns the transition model of the MDP. However, since the policy updates are based solely on an estimated model, it may not capture the true model and may learn a sub-optimal policy. Moreover, this approach will translate poorly into an active RL setting because of how long it might take for the collected statistics to reflect the true nature of the environment. 

\textbf{Model-free} methods rely heavily on the Law of Large Numbers to update the utility values without ever learning the model. This approach provides both computational and memory advantages because the state transitions and model does not need to be kept in memory and updated after each step in the environment. Temporal Difference (TD) learning \cite{sutton1988learning} is one of the most prominent model-free approaches that uses the estimated value of the next encountered state to update the value of the current one. Whenever a transition from state $s$ to state $s'$ occurs, the utility value of state $s$ is updates as follows:

\begin{equation}\label{eq:td}
U^{\pi}(s) \leftarrow U^{\pi}(s) + \alpha(R(s) + \gamma U^{\pi}(s') - U^{\pi}(s))
\end{equation}

The added parameter $\alpha\,\epsilon\,[0,1]$  is the learning rate, that value of which ranges from The average value of $U^{\pi}(s)$ will converge after a large number of episodes, taking longer than ADP but at a lower cost and with added flexibility of application to domains where the model cannot be learned  \cite{sutton1998reinforcement}. TD learning can be seen as a first approximation of ADP because it does not take into account all future states during each update (Recall the summation over all possible $s'$ in \eq{eq:iterative_bellman}). This means that, given a good exploration strategy and enough iteration, TD methods surpass ADP in active RL. In both these approaches, updating value estimates is done using other estimates; a process known as \textbf{bootstrapping}. A major advance in model-free RL came in the form of Q-learning \cite{watkins1989learning}, where an agent would learn the function $Q(s,a)$ that represents the quality of taking action $a$ in state $s$. This is much more practical than methods based on utility because the policy can be derived directly from the Q-function. 

The RL algorithms implemented in this work rely heavily on these ideas, and will be further elaborated on in \sect{sec:experiments}.

\subsection{Drawbacks of bootstrapping rewards}

While methods that use bootstrapping can provide guarantees for a solution (even an optimal one), it is conditioned on infinite exploration and assumes that a reward signal will be encountered in the environment \cite{sutton1998reinforcement}. In environments where rewards are very sparse, delayed in time compared to when an action was taken, or when multiple actions need to be executed in a specific order, bootstrapping cannot be guaranteed to succeed in a timely manner. Of course, infinite exploration and learning using $\epsilon-greedy$ means that successful strategies will be learned at some point given enough data \cite{watkins1989learning}, which is not practical in cases where simulations are computationally expensive or time consuming.

\subsection{Q-Learning}

Q-learning is a fairly flexible approach that can be applied to a host of different problems. Q-learning is a TD method, with an update equation identical to \eq{eq:td}: 
\begin{equation}\label{eq:q_learning}
    Q(s,a) \leftarrow{} Q(s,a) + \alpha(R(s) + \gamma max_{a'} Q(s',a') - Q(s,a))
\end{equation}

This algorithm operates in an off-policy manner, meaning that the target value used in the TD update in \eq{eq:q_learning} is the best possible future value that can be achieved in the next state, $max_{a'} Q(s', a')$, and not the value of the action actually taken. Standard Q-learning only looks ahead one-step ahead to update the value of the current state, making no use of longer sequences of steps to update multiple values. However, Q-learning is guaranteed to converge given infinite exploration and learning making it a good baseline for comparison. 

\subsection{SARSA($\lambda$)}

SARSA is another model-free algorithm that closely mirrors Q-learning, with a slight modification. SARSA is on-policy, so the value update is not carried out for the current state, until an actual action is taken in the next state. This approach can lead to faster convergence but is highly dependent on the policy used. Moreover, it is brittle when compared to Q-learning, which is more flexible when applied to different types of environments and robust across different hyperparameters and policies. The $\lambda$ portion of the title refers to the eligibility traces mechanism \cite{sutton1998reinforcement} that allows for multi-step look-ahead when updating the current state's value. Effectively, this maintains a memory of the sequence of state-action pairs that contribute to the currently encountered rewards and updates the entire sequence in the q-table.

\begin{figure}[t]
       \centering
       \captionsetup{justification=centering}
       \includegraphics[scale=0.6]{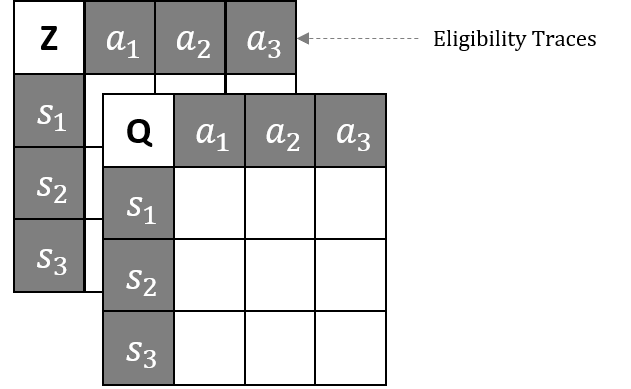}
       \caption[Illustration of SARSA($\lambda$)] {Illustration of SARSA($\lambda$): Uses an additional table to store the eligibility traces of state-action pairs in order execute multi-step updates, as opposed to the one-step updates in our Q-Learning implementation}
       \label{fig:sarsa}
\end{figure}

This is done by maintaining a separate table, refer to \fig{fig:sarsa}, where the eligibility of states for an update is maintained. Whenever a state-action pair is encountered, one is added to the corresponding table entry, and after each update, the entire table is multiplied by $\lambda$ (a hyperparameter, $\lambda\,\epsilon\,[0,1]$) which represents the decay of eligibility for each state-action pair for the next update. Updates occur as follows:

$$\Delta = R(s) + \gamma Q(s',a') - Q(s,a)$$
\begin{equation}
    Q \leftarrow{} Q + \alpha * \Delta * Z
\end{equation}

Note the lack of a $max_{a'}$ operator, which makes it on-policy. Also note the fact that the entire Q-table is updated at each step which does lead to increased computation when compared to Q-learning, especially with a high $\lambda$. Setting $\lambda = 0$ makes the algorithm one-step, as described in 3.1.1. 

\subsection{Dyna-Q}

The algorithms presented thus far do not use a model of the environment to learn. Using one however, would facilitate planning via simulated experience. One of the earliest methods to do so came with the Dyna architecture \cite{sutton1991dyna}. 

The model is built by adding newly encountered transitions to memory in a table of states and actions. The transition is stored as a tuple $(r,s')$ where $r$ is the reward received from taking the chosen action in the current state, and $s'$ is the next state the agent ends up in. The learning process is the same as Q-learning, with added planning steps. After each transition and Q-table update, previously encountered state-action pairs are sampled from the model and the stored tuple is used to execute the Q-learning update again on the sampled state-action pair. The added planning steps increase the speed of convergence.

\begin{figure}[t]
       \centering
       \captionsetup{justification=centering}
       \includegraphics[scale=0.35]{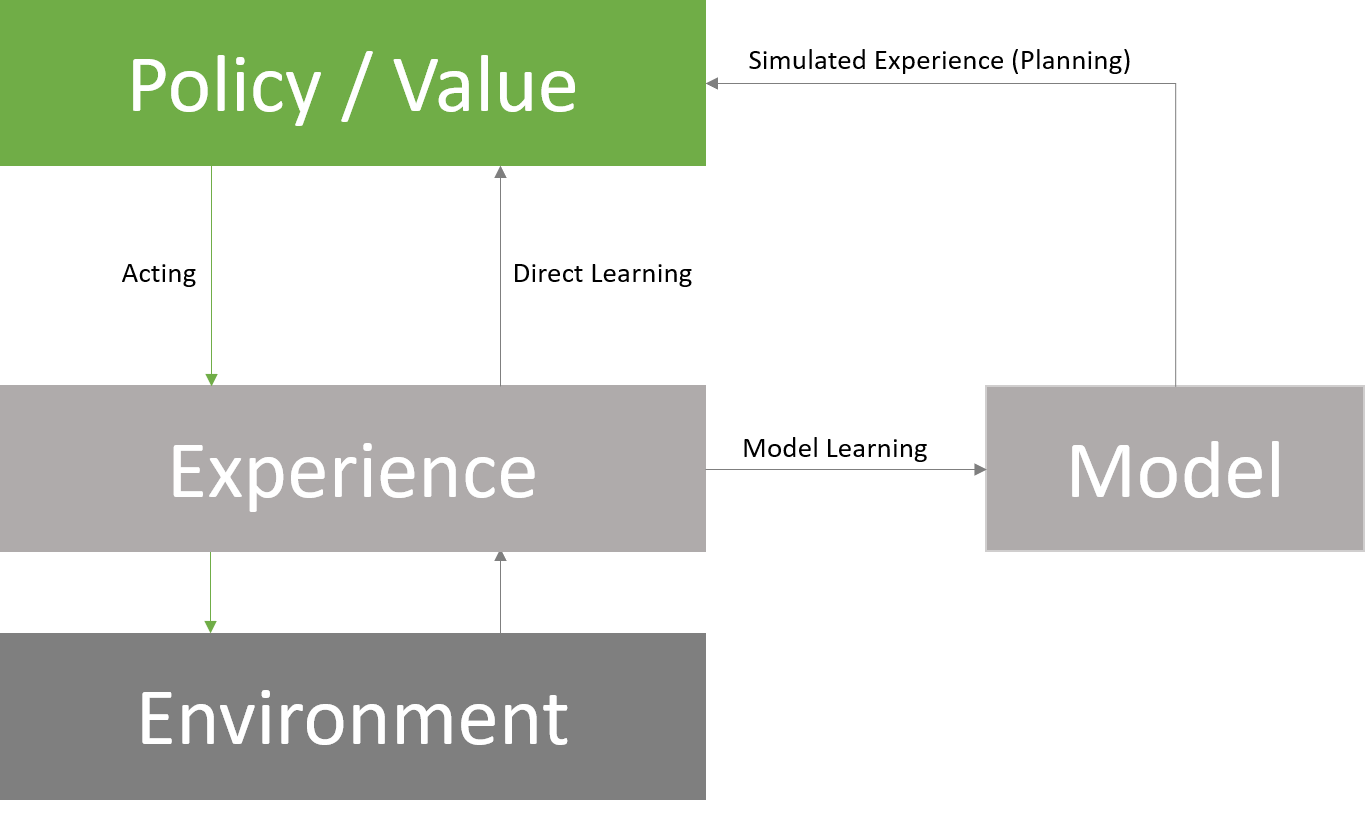}
       \caption{Dyna-Q Architecture: Incorporates a model of the environment to facilitate planning, which is done by sampling the stored transitions at every step and executing the same update as Q-Learning}
       \label{fig:dyna}
\end{figure}

The original implementation presented in \cite{sutton1991dyna} assumes that the environment is deterministic, which means that taking action $a$ in state $s$ will \textbf{always} result in reward and next state $(r,s')$. This is not a completely accurate representation of most environments encountered in everyday life, where a multitude of unconsidered factors can affect the consequences of agent actions, whereas simulated environments tend to be rigorously controlled. Moreover, this type of model will fail if the transitions are not the same every time. So in order to try making a more robust model that can handle stochastic environments, we implement both the original Dyna-Q in addition to a modified version that builds a model of the environment in a similar fashion to ADP, called \textit{Stochastic Dyna-Q}. This is done by counting the number of times state-action pairs are encountered, in addition to the number of times the specific state-action pair leads to state $s'$. Using this gathered information, we can compute the expected reward of taking a certain action in a state and use that for the planning portion of the algorithm. Planning differs significantly in this case because the Q-table entry being updated must reflect the new probabilities across all possible next states meaning that the target value of the TD update will be:

\begin{equation}
    \sum\limits_{s'} P(s'|s,a)\;max_{a'}Q(s',a').
\end{equation}



\section{Dyna-T}\label{sec:approach}

While a stochastic model is more robust, it is affected by the same problems as ADP, namely, the estimated model might take too long to accurately reflect the environment, which significantly slows down learning or even deteriorate it. To remedy this problem, we utilize the concept of Upper Confidence Bounds applied to Trees \cite{kocsis2006bandit}. This technique is used to improve Monte Carlo Tree Search (MCTS) methods by computing a statistical interval whose upper bound would represent the potential reward to be gained, while also emphasizing the exploration of new state-action pairs. The algorithm computes the upper bounds associated with a state-action pair with minimal increase in memory cost, but with a slightly increased computational one. The upper bounds are computed in a Dyna agent as follows:

\begin{equation}
    Q_{uct}(s,a) = Q(s, a) + c\sqrt{\frac{\ln n_s}{n_{s,a}}}
\end{equation}

Where $c$ is an experimentally determined exploration parameter, $n_s$ is the number of times state $s$ has been encountered, and $n_{s,a}$ is the number of times action $a$ was taken in state $s$.
An important fact to note is that this is not a modification to learning, but an improvement to the policy. Dyna-T does not use an $\epsilon-greedy$ policy like the rest of the algorithms presented so far, but chooses the action with the highest upper bound, which is initialized to infinity for unexplored state-action pairs to encourage exploratory behavior. This leads to an increase in the chances of encountering a reward in the environment via more directed exploration when compared to occasionally choosing random actions.


\section{Test Environments}

In this section we will introduce the environments used for our evaluation. We focus on deterministic or stochastic single-agent environments with fully observable states, discrete states and actions. All the environments are available in the OpenAI Gym toolkit \cite{1606.01540}.

\begin{figure*}[t]
    \centering
    \begin{subfigure}[b]{0.49\textwidth}
            \centering
            \includegraphics[width=\textwidth]{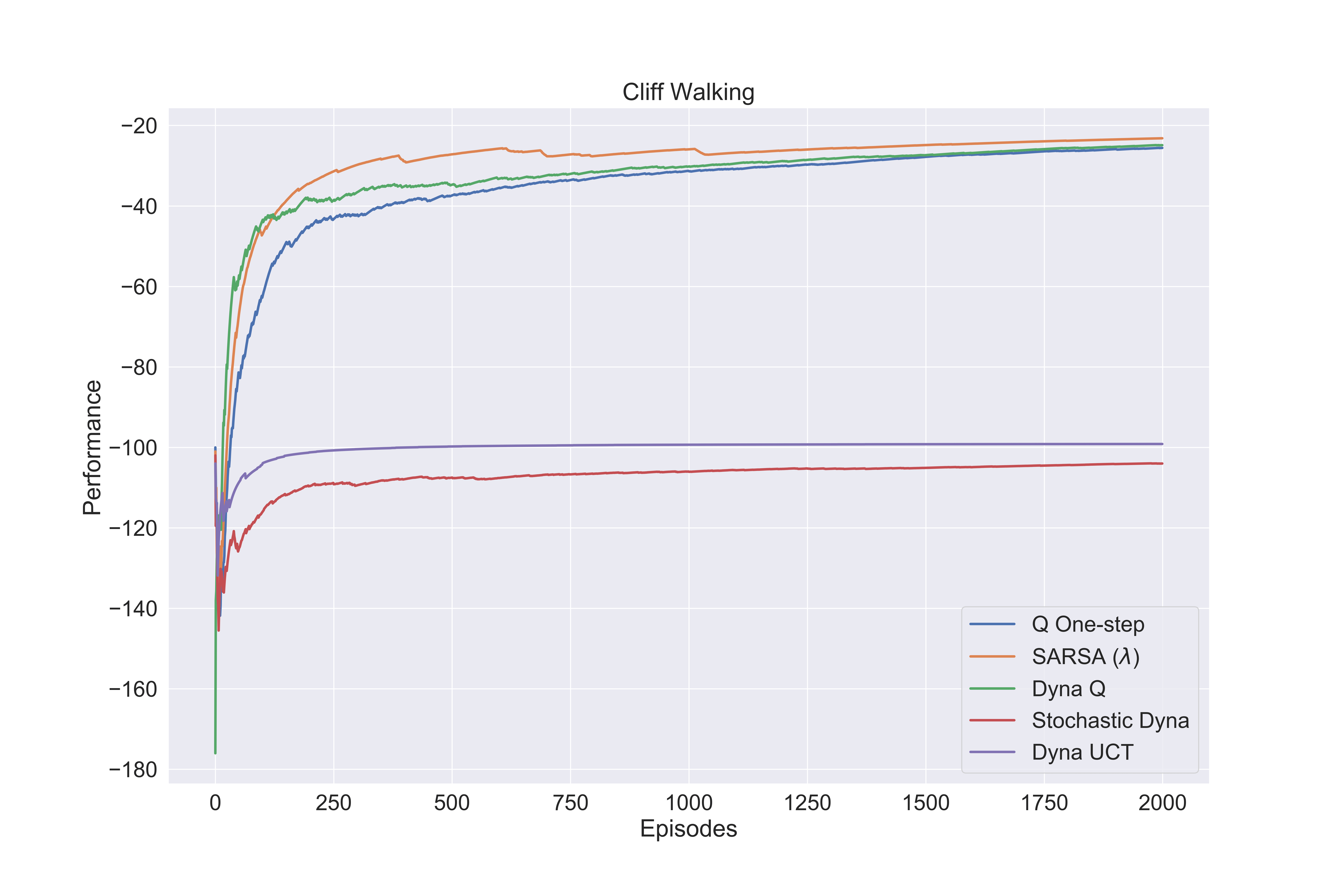}
    \caption{\label{fig:cw_rewards}}
    \end{subfigure}
\begin{subfigure}[b]{0.49\textwidth}
            \centering
            \includegraphics[width=\textwidth]{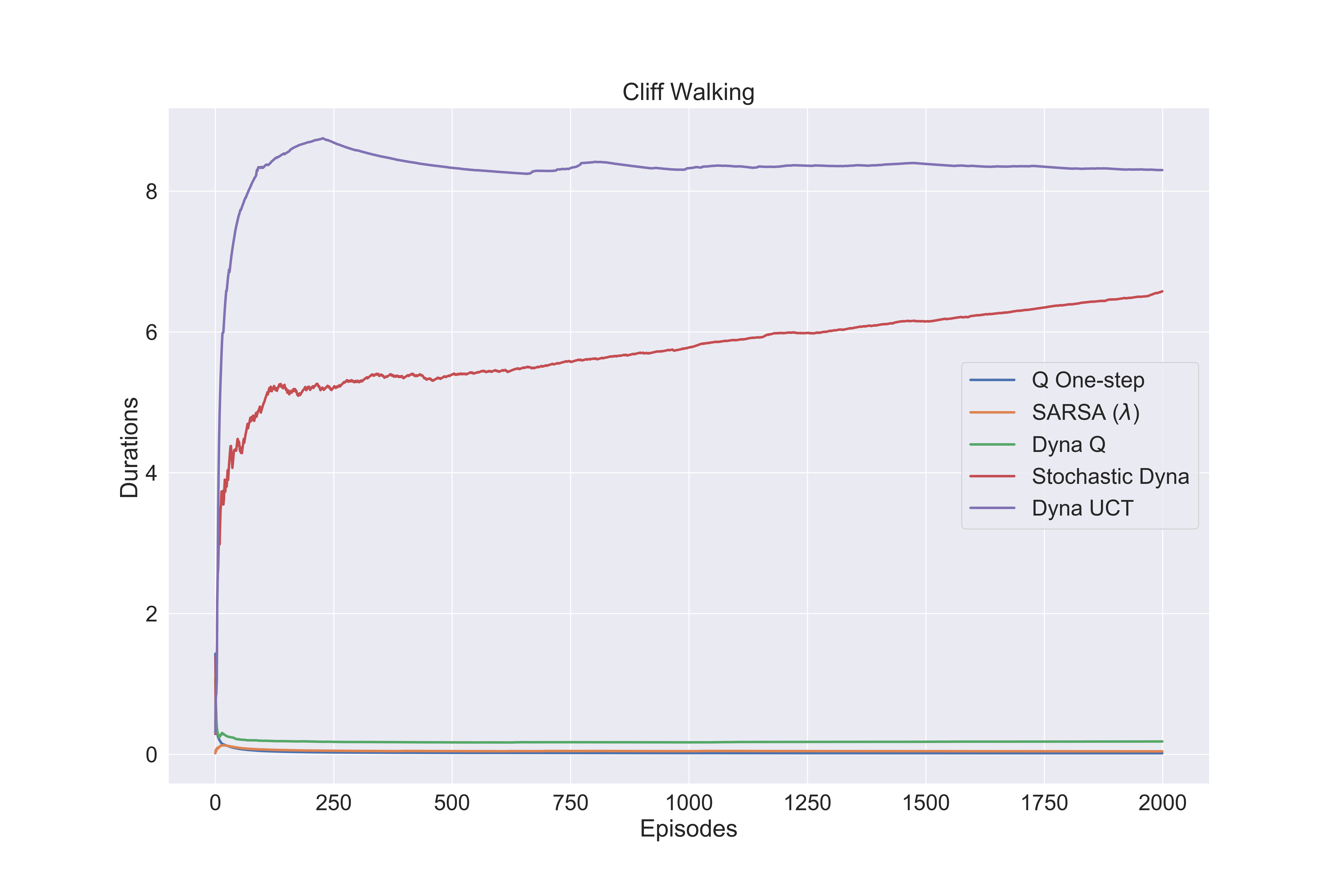}
     \caption{\label{fig:cw_duration}}
    \end{subfigure}
    \caption{CliffWalking Results: Stochastic Dyna-Q and Dyna-T fail to converge as quickly as the other algorithms.}
    \label{fig:cliffwalking_results}
\end{figure*}
\begin{figure*}[t]
    \centering
    \begin{subfigure}[b]{0.49\textwidth}
            \centering
            \includegraphics[width=\textwidth]{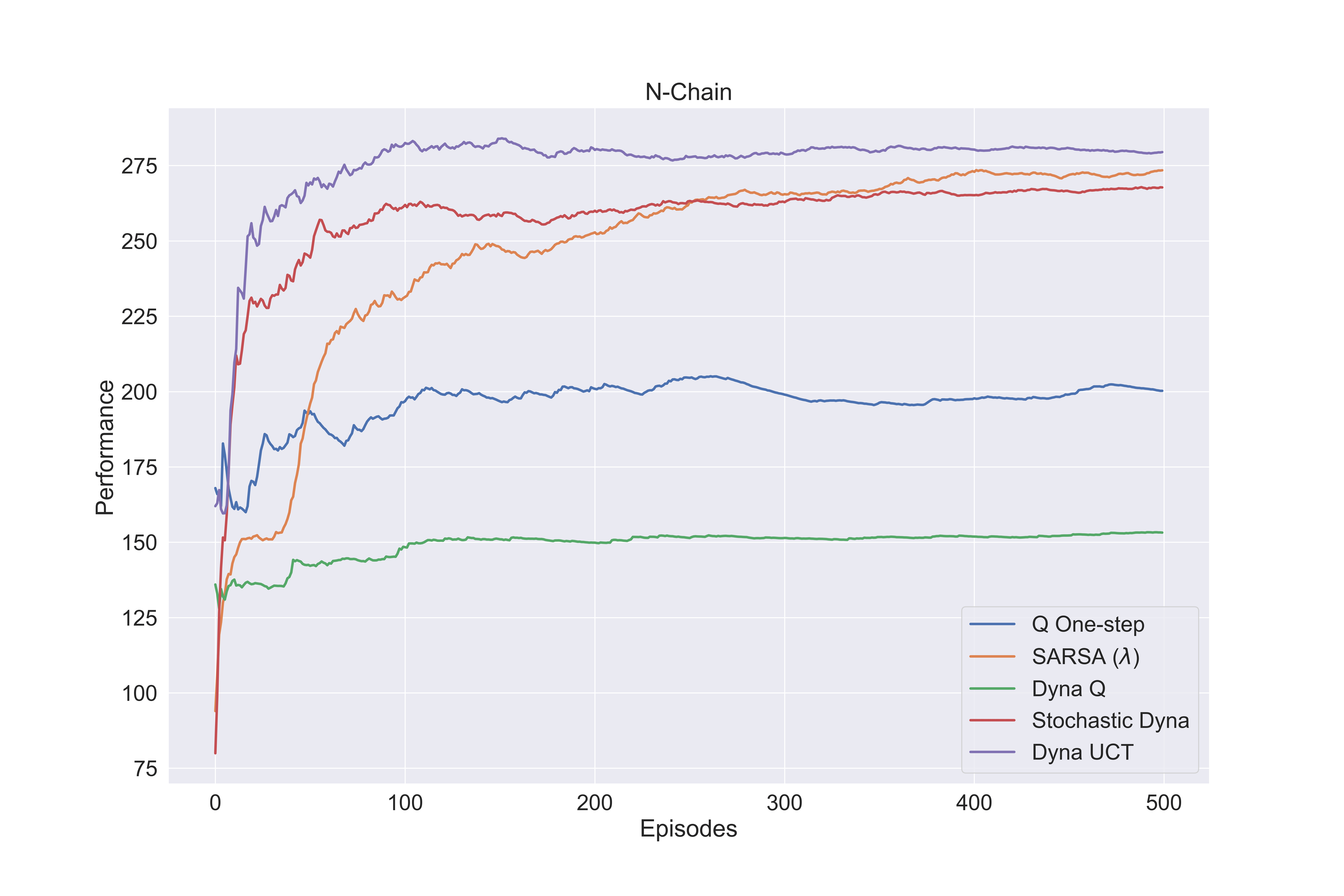}
    \caption{\label{fig:nc_rewards}}
    \end{subfigure}
\begin{subfigure}[b]{0.49\textwidth}
            \centering
            \includegraphics[width=\textwidth]{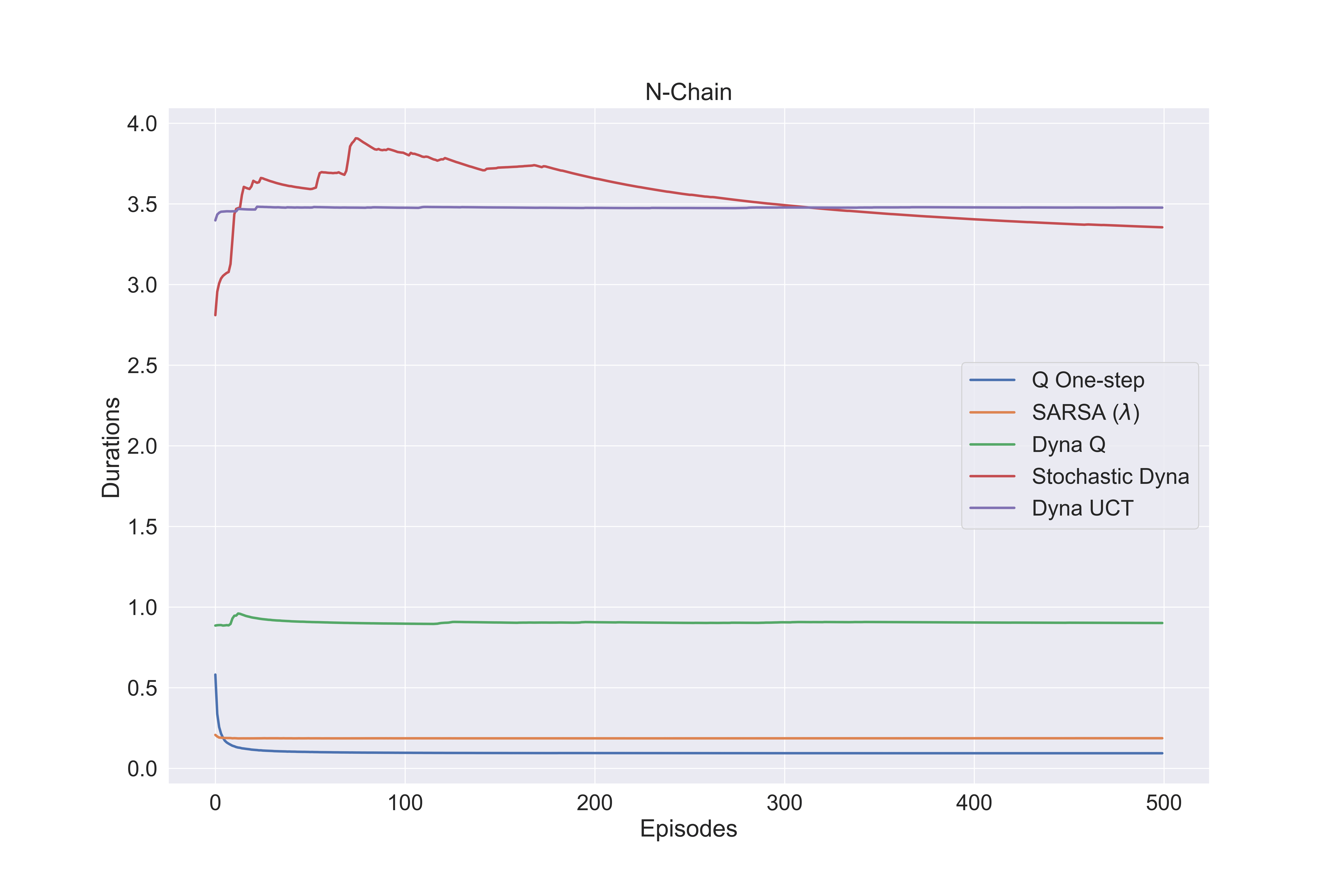}
     \caption{\label{fig:nc_duration}}
    \end{subfigure}
    \caption{NChain Results: Dyna-T converges on an optimal policy much quicker in the stochastic environment, besting the rest of the algorithms tested.}
    \label{fig:nchain_results}
\end{figure*}
\begin{figure*}[t]
    \centering
    \begin{subfigure}[b]{0.49\textwidth}
            \centering
            \includegraphics[width=\textwidth]{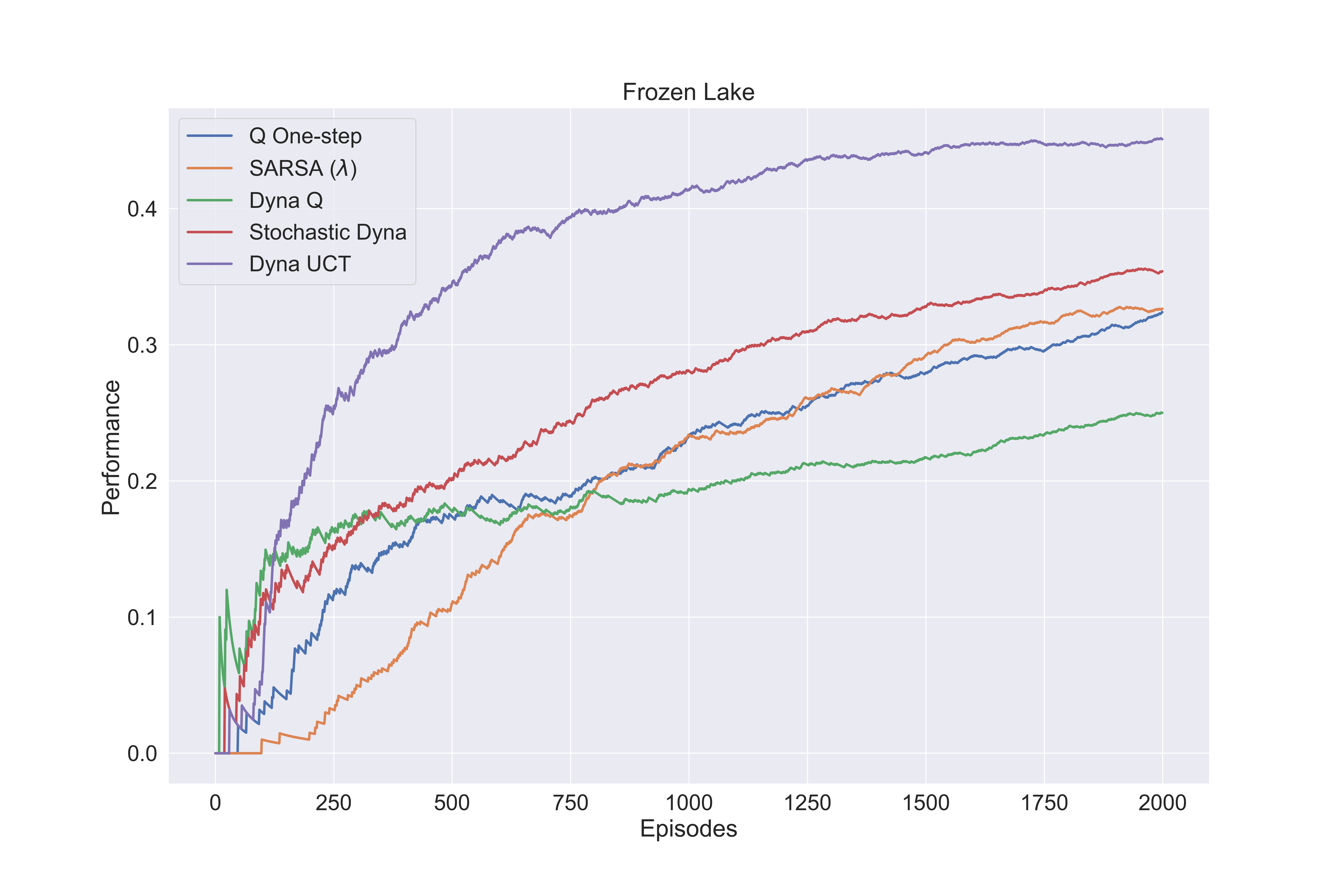}
    \caption{\label{fig:fl_rewards}}
    \end{subfigure}
\begin{subfigure}[b]{0.49\textwidth}
            \centering
            \includegraphics[width=\textwidth]{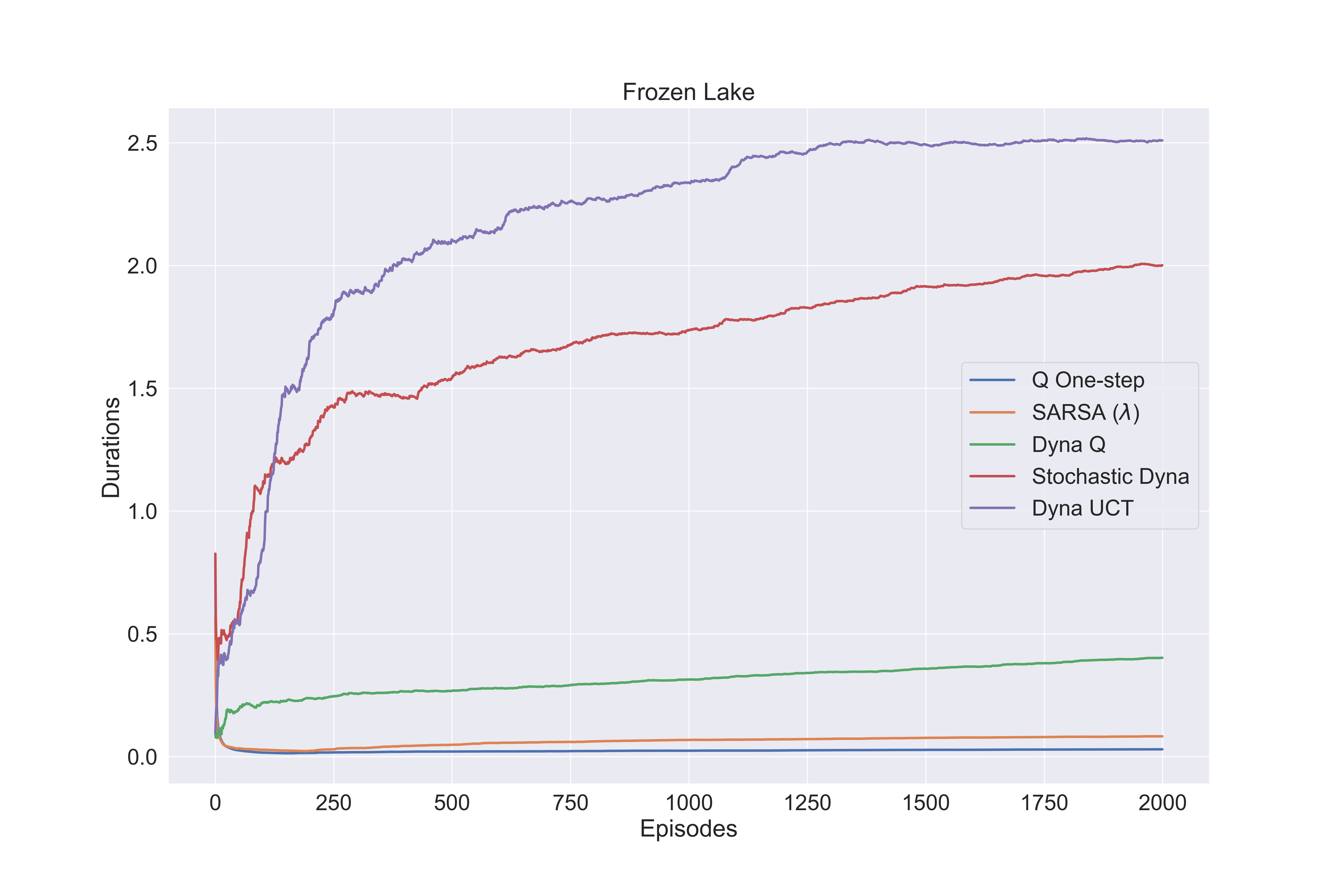}
     \caption{\label{fig:fl_duration}}
    \end{subfigure}
    \caption{FrozenLake Results: Dyna-T shows significantly better results on the most complex of the three environments presented, indicating a more robust action selection strategy.}
    \label{fig:frozenlake_results}
\end{figure*}

\subsection{Cliff Walking}

\begin{figure}[t]
       \centering
       \captionsetup{justification=centering}
       \includegraphics[scale=0.9]{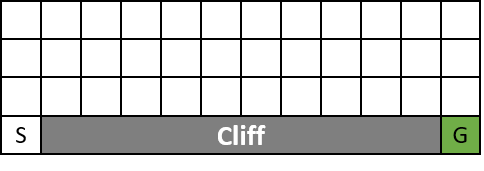}
       \caption[The CliffWalking Environment] {The CliffWalking Environment: An agent must get from the start to the goal in the shortest amount of steps since each additional step provides -1 as a reward. The agent loses if it steps into the cliff.}
\end{figure}

The first environment considered is a deterministic one with a dense reward distribution, presented in \cite{sutton1998reinforcement}. The agent must move from position \textbf{S} to \textbf{G}, and receives a -1 reward at every white square. If the agent falls in the cliff, it receives a reward of -100 and the episode is terminated. The goal square gives a reward of 0 and also terminates the episode. The agent's available actions are Up, Down, Left and Right that move it one square in the desired direction. The best performers are the ones that find the shortest path to the goal, and minimize the negative reward.

\subsection{N-Chain}

\begin{figure}[t]
       \centering
       \captionsetup{justification=centering}
       \includegraphics[scale=0.9]{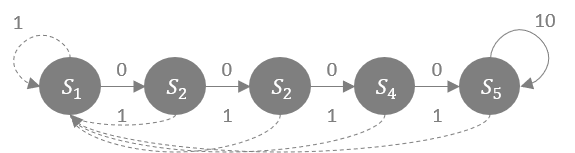}
       \caption[The NChain Environment] {The NChain Environment: The agent can only move forward and back, and the optimal agent ignores the small reward of moving back and chooses to go forward, despite the probability of "slipping" and taking the opposite action}
       \label{fig:nchain}
\end{figure}

The second environment is called N-Chain. There are only five states and two actions, forward and back, and a small probability that the action chosen is not the one actually taken in the environment. The agent starts in state \textbf{$S_1$}. Moving forward yields a reward of 0 except in \textbf{$S_5$} where the reward for moving forward is 10. Moving back sends the agent to state \textbf{$S_1$} and gives a reward of 1. Forward is represented by solid lines and backwards with a dashed line, as shown in \fig{fig:nchain}. Optimal behavior in this environment means to forgo the immediate rewards gained by moving back and continuously go forward to seek the larger reward available in the last state.

\subsection{Frozen Lake}

\begin{figure}[t]
       \centering
       \captionsetup{justification=centering}
       \includegraphics[scale=1]{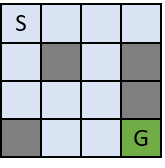}
       \caption[The FrozenLake Environment] {The FrozenLake Environment: The agent must navigate from start to finish, the grey squares are holes that terminate the episode with a reward of 0, while the blue squares are frozen and can cause an agent to slip in a random direction. The only positive reward available is 1, if the agent reaches the goal.}
       \label{frozenlake}
\end{figure}

The last environment combines elements from both of theones  presented  above and it is shown in \fig{fig:frozenlake}. The agent must navigate from \textbf{S} to \textbf{G}, and avoid holes, the grey squares, which terminate the episode with a reward of $0$. The blue squares are frozen, meaning the agent can slip in a random direction whenever it takes an action while it's on a blue square. The available actions are Up, Down, Left, and Right. The rewards are sparse when compared to the previous environments because the agent only receives a reward of $1$ if it manages to reach the goal state. Due to the nature of the environment, for an algorithm to solve this task it must achieve an average score of 0.74, at least, over $100$ episodes. 

\section{Experiments}\label{sec:experiments}

In this work, we assess the performance of Dyna-T against Q-Learning, SARSA($\lambda$), Dyna-Q, and Stochastic Dyna-Q using as a benchmarking platform the OpenAI Gym toolkit \cite{1606.01540}. Our chosen experimental scenarios are Cliff Walking, Frozen Lake, and NChain.

Q-Learning, SARSA($\lambda$), Dyna-Q, Stochastic Dyna-Q, and Dyna-UCT were implemented and run on each of the three chosen environments. CliffWalking and FrozenLake were tested for 2000 episodes, while NChain was tested for 500. Their behavior with different hyperparameters was observed using parameter evaluation scripts, that test each algorithm with different parameter values, such as learning rate or planning steps. The primary measure of performance was the reward collected in the environment, we also measure the duration across iterations of each algorithm for further insight into their practicality.

\section{Results \& Discussion}

The algorithms all behave as expected in the deterministic CliffWalking setting \fig{fig:cliffwalking_results}. SARSA($\lambda$) and Dyna-Q converge faster than one-step Q-learning, while Stochastic Dyna-Q and Dyna-T fail to learn the appropriate environment model quickly enough. Moreover, due to the added computational cost of storing transitions and updating expected rewards at every step Stochastic Dyna-Q and Dyna-T take a significantly longer time. This is also partially due to the use of the \textbf{Pandas} package to handle the tabular representation, storage, and retrieval operations of the algorithms. Pandas tends to be slower than a numerical computation package like NumPy, but does provide some advantages such as multi-indexing which was very helpful in the implementation. 

The balance shifts once the algorithms are tested in a stochastic environment. In NChain, as shown in \fig{fig:nchain_results}, one-step Q-learning and Dyna-Q fail to converge on an optimal solution within the allotted timeframe while SARSA($\lambda$), Stochastic Dyna-Q and Dyna-T perform much better. Dyna-T converges on the best solution which is in line with our expectations.The time difference is apparent, but less extreme due to the smaller state space and episode length. Note the particularly poor performance of Dyna-Q, because it assumes a deterministic environment, meaning that its model will store transitions that are not guaranteed to occur every time which can inhibit learning. 

FrozenLake is where we really start seeing the benefits of UCT, shown in \fig{fig:frozenlake_results}. Dyna-T latches onto solutions much quicker than the rest of the algorithms. While all of them are guaranteed to converge, Dyna-T will find the appropriate solutions faster despite taking no random actions to explore the environment, it is completely driven by the upper bounds computation.

\section{Conclusion and Future Work} 
\label{sec:conclusion}

In this paper, we presented a novel algorithm for on-line model-based reinforcement learning, called Dyna-T. We combined the benefits of two popular technique: i) Dyna-Q and ii) UCT. The former is a hybrid RL approach that combines planning, acting, model-learning, and direct-RL in a unique framework, all occurring continually at each iteration of the algorithm. However, the planning process can be computationally expensive and a drawback for this approach. On the other hand, UCT is a state-of-the-art sampling-based method for exploring large trees in an efficient way. We demonstrated that our Dyna-T is efficient in exploring the simulated experience generated by the Dyna-Q model leading to a more efficient action selection during the on-line learning. By interacting with the environment with better actions, our model converges quicker to the optimal solution than several state-of-the-art methods in two testbed environments from Open AI. We also compared the performance of our propose model in a deterministic scenario. Although Dyna-Q is deterministic by nature, the combination with the UCT degraded its performance in such scenario. Future work will focus on testing in larger environments and to compare it against deep-learning techniques.

%

\addtolength{\textheight}{-12cm}   


\bibliographystyle{plain}
\bibliography{ieee}

\end{document}